\documentclass{article}
\usepackage{amsmath,epsfig}
\usepackage[preprint]{spconfa4}
\usepackage{xcolor}
\usepackage{cite}
\usepackage{times}
\usepackage{graphicx}
\usepackage{amssymb}
\usepackage{url}
\usepackage{booktabs}
\usepackage{makecell}
\usepackage{multirow}
\usepackage[linesnumbered,lined,boxed,commentsnumbered,ruled]{algorithm2e}
\usepackage{rotating}

\usepackage{tabularx}
\usepackage{subfigure}
\usepackage{booktabs}
\usepackage{multirow}

\renewcommand{\paragraph}[1]{\noindent\textbf{#1}~~}

\let\OLDthebibliography\thebibliography
\renewcommand\thebibliography[1]{
  \OLDthebibliography{#1}
  \setlength{\parskip}{0pt}
  \setlength{\itemsep}{0pt plus 0.3ex}
}

\begin{document}\sloppy

\def\x{{\mathbf x}}
\def\L{{\cal L}}

\title{Domain Adaptive Semantic Segmentation via Regional Contrastive Consistency Regularization}
%

\name{Qianyu Zhou, Chuyun Zhuang, Ran Yi$^{1}$, Xuequan Lu$^{2 \dagger}$ and Lizhuang Ma$^{1 \dagger}$\thanks{$^{\dagger}$Corresponding authors.}}
\address{$^{1}$ Shanghai Jiao Tong University $^{2}$Deakin University \\ \{zhouqianyu,fallen,ranyi\}@sjtu.edu.cn,  xuequan.lu@deakin.edu.au, ma-lz@cs.sjtu.edu.cn}

\maketitle

\begin{abstract}
Unsupervised domain adaptation (UDA) for semantic segmentation has been well-studied in recent years. However, most existing works largely neglect the local regional consistency across different domains, and are less robust to changes in outdoor environments. In this paper, we propose a novel and fully end-to-end trainable approach, called regional contrastive consistency regularization (RCCR) for domain adaptive semantic segmentation. Our core idea is to pull the similar regional features extracted from the same location of different images, \emph{i.e.,} the original image and augmented image, to be closer, and meanwhile push the features from the different locations of the two images to be separated. We propose a region-wise contrastive loss with two sampling strategies to realize effective regional consistency. Besides, we present momentum projection heads, where the teacher projection head is the exponential moving average of the student. Finally, a memory bank mechanism is designed to learn more robust and stable region-wise features under varying environments. Extensive experiments demonstrate that our approach outperforms the state-of-the-art methods.
\end{abstract}

\section{Introduction}

Semantic segmentation aims to assign a semantic class to each pixel for a given image. Despite the great success in recent years, most methods~\cite{chen2018deeplab} heavily require a large amount of training data, and labeling such pixel-wise data is extremely expensive, and time-consuming~\cite{cordts2016cityscapes}.  
A natural idea is using synthetic data~\cite{stephan2016gtav,ros2016synthia} to supervise the segmentation model instead of real data.
However, such data cannot fully match the real-world distributions to guarantee reliable performance due to the existing domain shifts. 
Thus, it is necessary to reduce the labeling cost and improve the generalization ability of the segmentation models under different distributions. 

To cope with this problem, unsupervised domain adaptation (UDA) for semantic segmentation has been recently explored.  This task aims to bridge domain gaps between the labeled source domain and the unlabeled target domain.
Many approaches perform the adaptation in input-level~\cite{BDL,LDR}, feature-level~\cite{FDA,CBST,CRST}, and output-level~\cite{AdaptSegNet,SIM,CLANv2}. 
However, most of them heavily depend on the adversarial objectives~\cite{AdaptSegNet,SIM,CLANv2}, offline self-training~\cite{BDL,DAST,CBST,CRST} and image translation~\cite{BDL,LDR}, which make the training process too complicated and hard to converge. 

\begin{figure}[t]

\centering
\includegraphics[scale=0.42]{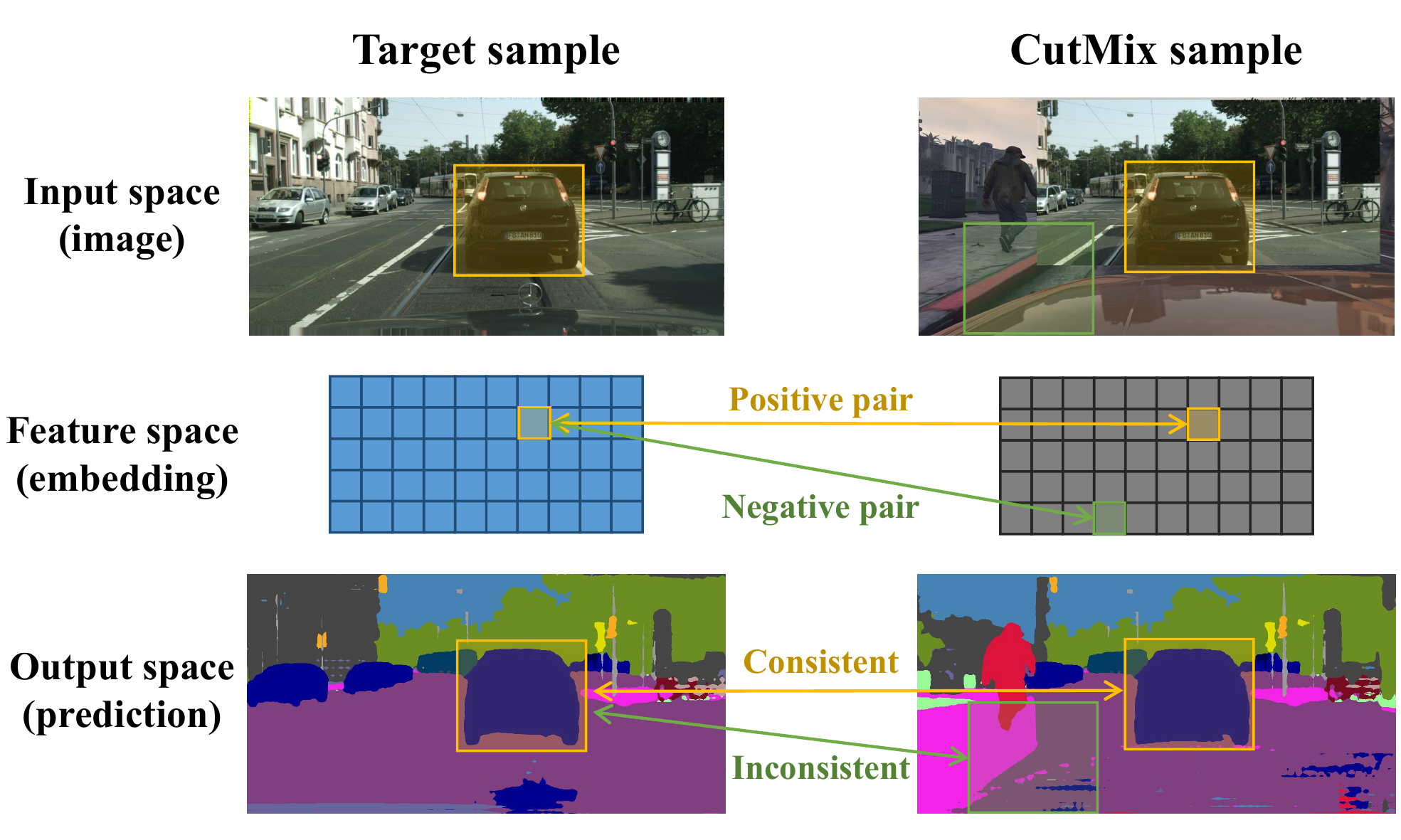}
\caption{ Previous domain adaptation methods overlook the regional consistency across different domains. To address this problem,
our key idea builds on region-level contrastive learning  by maximizing the inter-region differences and  minimizing intra-region disagreement. 1). On the output space, the predicted label should be invariant to cross-domain environmental augmentations, \emph{e.g.,} CutMix~\cite{french2019semi}. 2). On the feature space, we pull the similar regional embeddings extracted on the same location from the target image and mixed image to be closer, and push the dissimilar embeddings from the different locations of the two images to be separated. }
\label{fig0}
\end{figure}

Recently, consistency regularization~\cite{choi2019self,tranheden2020dacs}, also known as Mean Teacher, tackles this problem by employing the consistency constraint on the target prediction between the student model and the teacher model, respectively.  This kind of method usually performs the feature-level domain alignment between the student model and the teacher model with an online ensemble. The teacher model is an exponential moving average (EMA) of the student model, and then the teacher model could transfer the learned knowledge to the student.

Unfortunately, such methods usually employ an inconsistent penalty for the prediction map on the global level, while largely neglecting the region-wise consistency on the local level, \emph{i.e.,} some contextual object occurrence should be consistent regardless of the changes of outdoor environments. We observe that only capturing the pattern information from the global level is not powerful enough to enhance the feature-level representation in consistency regularization. If lacking such local regional information,  the segmentation result of objects will inevitably suffer from a non-marginal performance drop in the target domain. To prevent the model from abusing the contexts, we aim to make the learned representations more robust to the changing environments by exploring the regional consistency in a fine-grained manner.

Motivated by the above facts,  we propose a regional contrastive consistency regularization (RCCR) framework for domain adaptive semantic segmentation, which is fully end-to-end trainable. As shown in Fig. \ref{fig0}, our key idea builds on region-level contrastive learning by maximizing the inter-region differences and minimizing intra-region disagreement.
1) On the output label space, the predicted label should be invariant to cross-domain environmental augmentations, \emph{e.g.,} CutMix~\cite{french2019semi}. 2) On the feature space, we pull the similar regional features extracted from the same location of the target image and the augmented image to be closer, and simultaneously push the features from the different locations of the two images to be separated. 

Specifically, we first present the region-wise contrastive (RWC) loss between the target embeddings and the augmented embeddings, keeping the regional consistency in a fine-grained manner. Secondly, to fully utilize the spatial and local semantic cues, \emph{e.g.,} spatial layout and local context, we design momentum projection heads for aiding the region-level contrastive learning. The teacher projection head is the exponential moving average (EMA) of the student, and both of them are after the encoder architecture to produce the low-dimensional embeddings.
Besides, we present two sampling strategies for positive and negative samples,
avoiding treating every pixel-wise sample equally.
Finally, we introduce a memory bank mechanism to store the negative features created in the last few batches to learn more robust and stable region-wise features under varying environments. 


In a nutshell, our contributions are three-fold: 

$\bullet$ We propose a regional contrastive consistency regularization (RCCR) framework for domain adaptive semantic segmentation, which keeps the local regional consistency on the feature space and label space, respectively, under the cross-domain environmental augmentations. 
    
$\bullet$ We present a region-wise contrastive loss, sampling strategies, and momentum projection heads to realize effective regional consistency in domain adaptation. We also introduce a memory bank mechanism to learn more robust and stable region-wise features under varying environments. 
     
$\bullet$ We provide extensive experiments with analysis to demonstrate the state-of-the-art performance on two challenging benchmark datasets of domain adaptation.

\section{Related work}

\subsection{Domain Adaptive Semantic Segmentation}

Unsupervised domain adaptation (UDA) aims to learn a generalized model on the labeled source domain and the unlabeled target domain. This problem has been well-studied in semantic segmentation. 
Many recent approaches are proposed to reduce the domain gap between the source data and the target data on three different levels,  namely, the input-level~\cite{BDL,LTIR,LDR}, feature-level~\cite{CBST,CRST}, and output-level adaptation~\cite{AdaptSegNet,SIM,CLANv2}.  
However, most recent methods~\cite{BDL,FDA,SIM,LTIR} involve many sophisticated sub-components, \emph{e.g.,} computationally-involved adversarial objectives~\cite{AdaptSegNet,SIM,CLANv2}, offline self-training~\cite{BDL,DAST,CBST,CRST} and image translation models~\cite{BDL,LTIR,LDR}, which are hard to converge, and cannot be trained in an end-to-end manner. Besides, most methods are  less  robust  to changing of outdoor environments.
In contrast, our method is effective and does not require any fine-tuning, and we explore the regional consistency across domains.  On top of \cite{tranheden2020dacs}, we realize fully end-to-end training.

\begin{figure*}[t]
\centering
\includegraphics[width=1.0\textwidth]{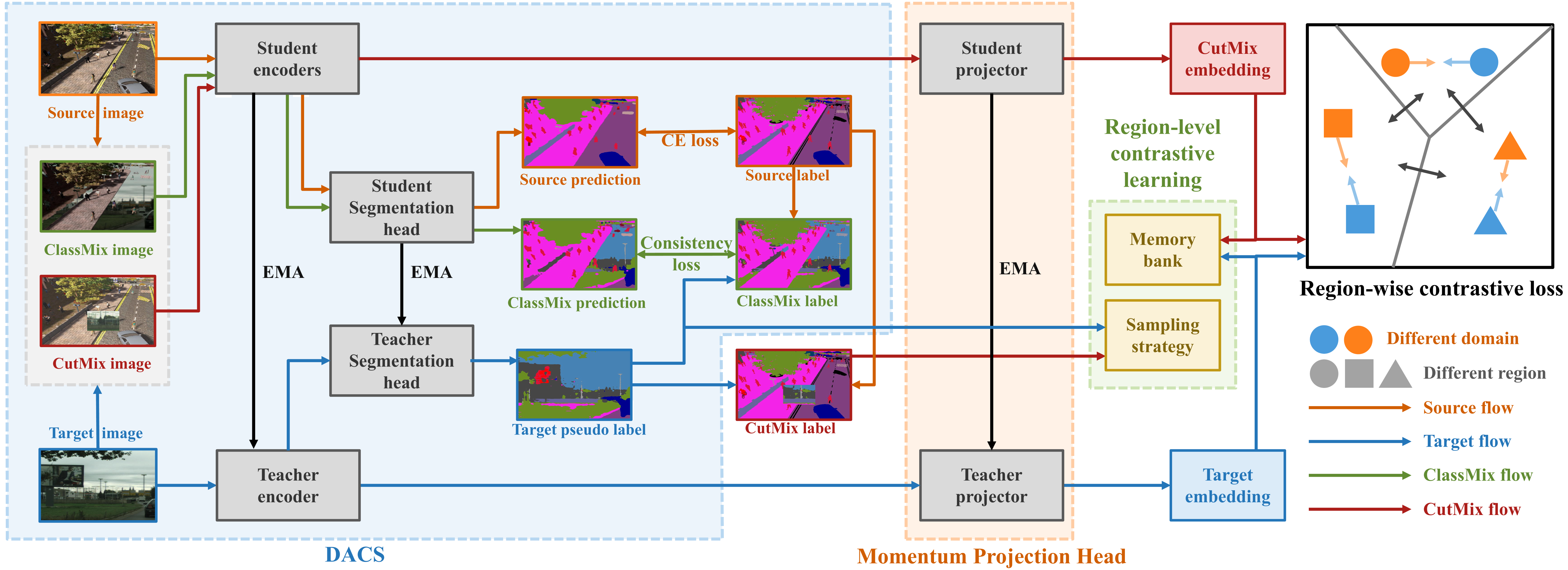}
\caption{Overview of the regional contrastive consistency regularization (RCCR) architecture. Firstly, to produce cross-domain environmental changes, we cut a region from the target and paste it onto the source image to generate CutMix images.
Then, we design momentum projection heads, namely, the student and teacher projector, to extract CutMix embeddings and target embeddings, respectively. Given these embeddings, we perform the region-level contrastive learning. Concretely, we compute the proposed region-wise contrastive (RWC) loss  by maximizing  the  inter-region  differences  and  minimizing intra-region disagreement. Moreover, we introduce sampling strategies and memory banks to further enhance our contrastive learning paradigm.
}
\label{fig1}
\end{figure*}

\subsection{Contrastive Learning}

Great progress in contrastive learning has been achieved  by encouraging the positive pairs to get closer and pulling the negative pairs apart. For semantic segmentation tasks,  
the definition of positive pairs and negative pairs can be various to fit the dense pixel prediction requirements. ~\cite{wang2021exploring} treated the same category samples as the positive pairs and others as the negative pairs. ~\cite{liu2021domain} divided the positive pairs and negative pairs according to the label distribution similarity between different patches. There are also some works that investigated the contrastive learning methods ~\cite{lai2021semi} in Semi-Supervised Semantic Segmentation (SSS). \textit{Our method differs from these methods in several aspects}. Firstly, we tackle a more complicated task UDA rather than SSS, where the domain shifts exist between the source and the target domain. Secondly, most of them only consider category-wise contrastive learning patterns while largely neglecting the region-wise consistency across domains. In contrast, we attempt to keep the regional consistency in the feature space and label space.

\section{Method}

\subsection{Notations and Overview}
\label{sec:3.1}
In the UDA task, we have access to the source domain with labels, denoted as $D_{s}=\left\{\left(x_{s}, y_{s}\right) \mid x_{s} \subset\right.$ $\left.\mathbb{R}^{H \times W \times 3}, y_{s} \subset \mathbb{R}^{H \times W}, y_{s} \in[1, C]\right\}$, and the target domain without labels is $D_{t}=\left\{\left(x_{t}\right) \mid x_{t} \subset\right.$ $\left.\mathbb{R}^{H \times W \times 3}\right\}$. Our primary goal is to bridge the domain gap between $D_{s}$ and $D_{t}$.

Fig. \ref{fig1} describes the whole pipeline of the RCCR in the end-to-end training procedure. Our framework includes a Mean Teacher~\cite{choi2019self}, momentum projection heads, and region-wise contrastive learning. To produce cross-domain environmental changes, we cut a random region from the target and paste it onto the source image to generate CutMix image $x_{cut}$. Then, we design the student and teacher projection head~$F_{proj}$ to extract CutMix embeddings~$e^{cut}$ and target embeddings~$e^t$, respectively. Given these embeddings, we perform the region-level contrastive learning by maximizing  the  inter-region  differences  and  maximizing intra-region agreement. Moreover, we introduce sampling strategies and memory banks to further learn more robust and stable region-wise features under varying environments.  Following DACS~\cite{tranheden2020dacs}, we compute a Cross-Entropy loss $\mathcal{L}_{CE}$ for supervising the student model with source labels, and a consistency loss $\mathcal{L}_{cons}$ between the ClassMix prediction $p_{class}$ and the ClassMix label $y_{class}$.  ClassMix~\cite{tranheden2020dacs} is a kind of augmentation by pasting some selective classes of the source onto the target image.
In the inference phase, we only keep the teacher model.

\subsection{Regional Contrastive Consistency Regularization}
\label{sec:3.2}

\paragraph{Momentum Projection Head.}
As shown in Fig. \ref{fig1}, we design momentum projection heads, namely student projector and teacher projector, where the teacher projector is an exponential moving average (EMA) of the student projector, $\hat \theta_{proj}^{(t)}=\alpha\hat \theta_{proj}^{(t-1)}+(1-\alpha)\theta_{proj}^{(t)}$. The EMA of the model is more stable and more precise than the student projector, thus can provide high-quality embeddings for the contrastive learning.
The projection heads $F_{proj}$
behind the feature extractor $F_{enc}$ map the latent high-dimensional features $z \in \mathbb{R}^{h \times w \times D}$ of $F_{enc}$ to low ones $e=F_{\mathrm{proj}}(z) \in \mathbb{R}^{h \times w \times K}$, where channel number $K \textless D$. 
Instead of using the outputs $z$ of the feature extractor, the main intuition of the projection head is to prevent losing too many  spatial and local semantic cues in the adaptation. Such rich information, \emph{e.g.,}  spatial layout and local context, can be well used to aid the region-wise contrative learning.
We implement $F_{proj}$ with two consecutive convolutional layers and ReLU. 

\begin{table*}
\caption{Comparison results (mIoU) from GTAV to Cityscapes. }
\label{table:gtav}
\centering
\resizebox{\textwidth}{!}{%
\begin{tabular}{c|ccccccccccccccccccc|c}
\toprule
Method& \begin{turn}{90}road\end{turn} & \begin{turn}{90}sidewalk\end{turn} & \begin{turn}{90}building\end{turn} & \begin{turn}{90}wall\end{turn} & \begin{turn}{90}fence\end{turn} & \begin{turn}{90}pole\end{turn} & \begin{turn}{90}light\end{turn} & \begin{turn}{90}sign\end{turn} & \begin{turn}{90}vegetation\end{turn} & \begin{turn}{90}terrain\end{turn} & \begin{turn}{90}sky\end{turn} & \begin{turn}{90}person\end{turn} & \begin{turn}{90}rider\end{turn} & \begin{turn}{90}car\end{turn} & \begin{turn}{90}truck\end{turn} & \begin{turn}{90}bus\end{turn} & \begin{turn}{90}train\end{turn} & \begin{turn}{90}motocycle\end{turn} & \begin{turn}{90}bike\end{turn} & \begin{turn}{90}\textbf{mIoU$_{19}$}\end{turn}  \\ 
\midrule
Source Only  &63.3 &15.7 &59.4 &8.6 &15.2 &18.3 &26.9 &15.0 &80.5 &15.3 &73.0 &51.0 &17.7 &59.7 &28.2 &33.1 &3.5 &23.2 &16.7 &32.9  \\ 
\midrule
BDL~\cite{BDL} &91.0 &44.7 &84.2 &34.6 &27.6 &30.2 &36.0 &36.0 &85.0 &43.6 &83.0 &58.6 &31.6 &83.3 &35.3 &49.7 &3.3 &28.8 &35.6 &48.5 \\
SIM~\cite{SIM} &90.6 &44.7 &84.8 &34.3 &28.7 &31.6 &35.0 &37.6 &84.7 &43.3 &85.3 &57.0 &31.5 &83.8 &42.6 &48.5 &1.9 &30.4 &39.0 & 49.2 \\
FDA~\cite{FDA} &92.5 &53.3 &82.4 &26.5 &27.6 &36.4 &40.6 &38.9 &82.3 &39.8 &78.0 &62.6 &34.4 &84.9 &34.1 &53.1 &\textbf{16.9} &27.7 &46.4 &50.5 \\
WLabel~\cite{WLabel} &91.6 &47.4 &84.0 &30.4 &28.3 &31.4 &37.4 &35.4 &83.9 &38.3 &83.9 &61.2 &28.2 &83.7 &28.8 &41.3 &8.8 & 24.7 &46.4 &48.2 \\
FADA~\cite{FADA} &92.5 &47.5 &85.1 &37.6 &32.8 &33.4 &33.8 &18.4 &85.3 &37.7 &83.5 &63.2 &39.7 &87.5 &32.9 &47.8 &1.6 &34.9 &39.5 & 49.2 \\
LDR~\cite{LDR} &90.8 &41.4 &84.7 &35.1 &27.5 &31.2 &38.0 &32.8 &85.6 &42.1 &84.9 &59.6 &34.4 &85.0 &42.8 &52.7 &3.4 &30.9 &38.1 &49.5\\
CCM~\cite{CCM} &\textbf{93.5} &\textbf{57.6} &84.6 &39.3 &24.1 &25.2 &35.0 &17.3 &85.0 &40.6 &\textbf{86.5} &58.7 &28.7 &85.8 &\textbf{49.0} &56.4 &5.4 &31.9 &43.2 &49.9 \\
ASA~\cite{ASA}  &89.2 &27.8 &81.3 &25.3 &22.7 &28.7 &36.5 &19.6 &83.8 &31.4 &77.1 &59.2 &29.8 &84.3 &33.2 &45.6 &\textbf{16.9} &34.5 &30.8 &45.1 \\
CLAN~\cite{CLANv2}  &88.7 &35.5 &80.3 &27.5 &25.0 &29.3 &36.4 &28.1 &84.5 &37.0 &76.6 &58.4 &29.7 &81.2 &38.8 &40.9 &5.6 &32.9 &28.8 &45.5\\
DAST~\cite{DAST} &92.2 &49.0 &84.3 &36.5 &28.9 &33.9 &38.8 &28.4 &84.9 &41.6 &83.2 &60.0 &28.7 &87.2 &45.0 &45.3 &7.4 &33.8 &32.8 &49.6 \\
BiMaL~\cite{truong2021bimal}  & 91.2 & 39.6 & 82.7 & 29.4 & 25.2 & 29.6 & 34.3 & 25.5 & 85.4 & 44.0 & 80.8 & 59.7 & 30.4 & 86.6 & 38.5 & 47.6 & 1.2 & 34.0 & 36.8 & 47.3\\
\midrule
Ours &93.2 &54.7 &\textbf{86.7} &\textbf{42.1} &\textbf{34.9} &\textbf{37.9} &\textbf{44.4} &\textbf{42.8} &\textbf{87.5} &\textbf{51.2} &86.1 &\textbf{65.5} &\textbf{37.8
 } &\textbf{88.5} &47.2 &\textbf{62.2} &5.4 &\textbf{35.5} &\textbf{47.0} &\textbf{55.3}\\
\bottomrule
 \end{tabular}
 }
\end{table*}

\paragraph{Region-wise Contrastive Loss.}
The core idea of region-wise contrastive (RWC) loss is to keep the regional consistency on a fine-grained level by maximizing the inter-region differences and minimizing the intra-region disagreements.  In other words, we aim to pull the embedding on the same location of the overlap region between the CutMix embedding  $e^{cut}$ and the target embedding $e^{t}$ to be closer, and push the embeddings on other locations to be separated.  The proposed contrastive loss function $\mathcal{L}_{cont}$ is defined as: 
\begin{equation}
\begin{aligned}
&\mathcal{L}_{cont}=-\frac{1}{N} \sum_{i=1}^{N} \frac{1}{\left|\mathcal{P}_{i}\right|} \\
&\sum_{j \in \mathcal{P}_{i}} \log \frac{\exp \left(e_{i}^{t} \cdot e^{cut}_{j} / \tau\right)}{\exp \left(e_{i}^{t} \cdot e^{cut}_{j} / \tau\right)+\sum_{k \in \mathcal{N}_{i}} \exp \left(e_{i}^{t} \cdot e^{cut}_{k} / \tau\right)},
\end{aligned}
\end{equation}
where $e_j$ 
denotes the positive embeddings from the student and $e_k$ denotes the negative embeddings from the teacher and student.
$\mathcal{P}_{i}$ and $\mathcal{N}_{i}$ denotes the corresponding positive set and negative set on the $i$-th region. $N$ is the total number of regions and $\tau$ represents the temperature. On one hand, the discrepancy between the elements of the positive pair should be minimized, thus maximizing the intra-region agreements and performing intra-domain adaptation. On the other hand, the discrepancy between the elements of the negative pair should be constrained, thus maximizing the inter-region difference and performing inter-domain adaptation. RWC loss is an asymmetric contrastive learning since the teacher model tends to output higher confident prediction than the student due to EMA, and thus we only make $e^{cut}$ close to $e^{t}$. 

\paragraph{Sampling Strategies.}
To avoid treating every sample equally, we introduce two sampling strategies of positive and negative samples in region-level contrastive learning.
(1) For negative samples, we introduce random sampling and category-aware sampling strategies. Specifically, we first randomly select half samples from the original negatives. Secondly, we gather the negative embeddings that have the same label or pseudo label with $e^{cut}$. Note that we do not enlarge the embedding distance for the same category.
(2) For positive sampling, we set a threshold $\epsilon$ to select positive samples, for every $e^{cut}$, and only the prediction probability in the same location of corresponding target image larger than $\epsilon$ will be included in contrastive learning.

\paragraph{Memory Bank Mechanism.}
To further enhance the contrastive learning scheme, we develop a memory bank mechanism. To be specific, the embedding outputs produced in the last few batch images are also considered as negatives in the current iteration, and we construct a memory bank to save useful information:  the projection embedding and their corresponding label or pseudo label.   The introduced memory bank is driven by the motivation that the environments between different images may have similar distributions and can be semantically related due to the fact that they come from the same dataset. Therefore, this strategy can enable learning more robust and stable region-wise features under varying environments. Besides, it can also reduce the memory occupation and computational resources.   

\begin{table*}[t]
\caption{Comparison results (mIoU) from SYNTHIA to Cityscapes.}
\label{table:synthia}
\centering
\resizebox{1.0\textwidth}{!}{%
\begin{tabular}{c|cccccccccccccccc|c|c} 
\toprule
Method &  \begin{turn}{90}road\end{turn} & \begin{turn}{90}sidewalk\end{turn} & \begin{turn}{90}building\end{turn} &
\begin{turn}{90}wall$^{*}$\end{turn} &
\begin{turn}{90}fence$^{*}$\end{turn} &
\begin{turn}{90}pole$^{*}$\end{turn} &
\begin{turn}{90}light\end{turn} & \begin{turn}{90}sign\end{turn} & \begin{turn}{90}vegetation\end{turn} & \begin{turn}{90}sky\end{turn} & \begin{turn}{90}person\end{turn} & \begin{turn}{90}rider\end{turn} & \begin{turn}{90}car\end{turn} & \begin{turn}{90}bus\end{turn} & \begin{turn}{90}motocycle\end{turn} & \begin{turn}{90}bike\end{turn} &
\begin{turn}{90}\textbf{mIoU$_{16}$}\end{turn} & \begin{turn}{90}\textbf{mIoU$_{13}$}\end{turn}  \\ 
\midrule
Source Only & 19.6 & 12.8 & 70.4 &10.8 &0.1 &24.5 &5.9 &11.7 &67.4 &76.0 &52.1 &16.3 &59.4 &19.7 &14.6 &16.5 &29.9 &34.0\\
\midrule
BDL~\cite{BDL} &86.0 &46.7 &80.3 &-- &-- &-- &14.1 &11.6 &79.2 &81.3 &54.1 &27.9 &73.7 &42.2 &25.7 &45.3  &-- &51.4 \\
SIM~\cite{SIM}& 83.0 &44.0 &80.3 &-- &-- &-- & 17.1 &15.8 &80.5 &81.8 &59.9 &33.1 &70.2 &37.3 &28.5 &45.8  & -- &52.1 \\
FDA~\cite{FDA}& 79.3 &35.0 &73.2 &-- &-- &-- &19.9 &24.0 &61.7 &82.6 &61.4 &31.1 &83.9 &40.8 &\textbf{38.4} &51.1  & -- &52.5\\
WLabel~\cite{WLabel}&92.0 &53.5 &80.9 &11.4 &0.4 &21.8 &3.8 &6.0 &81.6 &84.4 &60.8 &24.4 &80.5 &39.0 &26.0 &41.7  &44.3 &51.9 \\
CCM~\cite{CCM}&79.6 &36.4 &80.6 &13.3 &0.3 &25.5 &22.4 &14.9 &81.8 &77.4 &56.8 &25.9 &80.7 &45.3 &29.9 &\textbf{52.0}  & 45.2 &52.9 \\
LDR~\cite{LDR}&85.1 &44.5 &81.0 &-- &-- &-- &16.4 &15.2 &80.1 &84.8 &59.4 &31.9 &73.2 &41.0 &32.6 &44.7 & -- &53.1 \\
CLAN~\cite{CLANv2} &82.7 &37.2 &81.5 &-- &-- &-- &17.1 &13.1 &81.2 &83.3 &55.5 &22.1 &76.6 &30.1 &23.5 &30.7 & -- &48.8\\
ASA~\cite{ASA} &91.2 &48.5 &80.4 &3.7 &0.3 &21.7 &5.5 &5.2 &79.5 &83.6 &56.4 &21.9 &80.3 &36.2 &20.0 &32.9 & 41.7 &49.3\\
DAST~\cite{DAST}&87.1 &44.5 &82.3 &10.7 &0.8 &29.9 &13.9 &13.1 &81.6 &86.0 &60.3 &25.1 &83.1 &40.1 &24.4 &40.5 & 45.2 &52.5\\
BiMaL~\cite{truong2021bimal} & \textbf{92.8} & 51.5 & 81.5 &10.2 &1.0 &30.4  & 17.6 & 15.9 & 82.4 & 84.6 & 55.9 & 22.3 & 85.7 & 44.5 & 24.6 & 38.8 & 46.2 &53.7 \\
\midrule
Ours &92.5 &\textbf{58.7} &\textbf{83.7} &\textbf{15.1} &\textbf{1.3} &\textbf{34.7}  &\textbf{26.6} &\textbf{27.1} &\textbf{82.6} &\textbf{87.3} &\textbf{66.0
} &\textbf{34.9} &\textbf{86.5} &\textbf{50.5} &23.6 &47.4  & \textbf{51.1} &\textbf{59.0}\\
\bottomrule
\end{tabular}}
\end{table*}

\section{Experiments}

\subsection{Experimental Setup}
\label{sec:4.1}
Following common UDA protocols~\cite{AdaptSegNet,SIM,BDL}, our experiments are conducted on two widely-used UDA benchmarks,  \emph{i.e.,} GTAV~\cite{stephan2016gtav} $ \rightarrow $ Cityscapes~\cite{cordts2016cityscapes} and SYNTHIA~\cite{ros2016synthia} $\rightarrow$ Cityscapes~\cite{cordts2016cityscapes}, and we adopt Deeplabv2~\cite{chen2018deeplab} framework with a ResNet101 backbone as our model. The backbone is pre-trained on ImageNet. 
Following \cite{tranheden2020dacs}, we also apply Color jittering and Gaussian blurring on the mixed images . 
The projection head $F_{proj}$ is constructed by two consecutive convolutions (2048 hidden layer channels and 128 output channels) with one intermediate ReLU layer.  The temperature $\tau$ is set to 0.1, and the positive threshold $\epsilon$ is set to 0.75 by default. 
We use $batch size=2$ for $250000$ iterations in all experiments. 

\subsection{Comparison to State-of-the-Art Methods}
\label{sec:4.2}
We compare our method with the state-of-the-art UDA methods on two common UDA benchmarks, \emph{i.e.,} GTAV~\cite{stephan2016gtav} $ \rightarrow $ Cityscapes~\cite{cordts2016cityscapes} in Table \ref{table:gtav} and the results of SYNTHIA~\cite{ros2016synthia} $\rightarrow$ Cityscapes~\cite{cordts2016cityscapes} in Table \ref{table:synthia}, respectively.  

As shown in these two tables, our proposed RCCR method outperforms the state-of-the-art approaches by $5\% \sim 6\%$ on two challenging tasks. It also surpasses the baseline (``Source Only'') by around $22\%$ and $25\%$, respectively. Specifically, we obtain a $55.3\%$ mIoU on GTAV~\cite{stephan2016gtav} and achieve the best per-class IoU performance for 14 classes among the total 19 classes. For SYNTHIA~\cite{ros2016synthia} dataset, we observe a $59.0\%$ mIoU of 13 classes and $51.1\%$ mIoU of 16 classes, and get the best per-class IoU in 13 classes among the total 16 classes. These results reveal the effectiveness of our RCCR among different classes, \emph{e.g.,} building, traffic light, traffic sign, person, rider, car, etc. 

	\begin{table}
	   \label{table:ablation}
		\centering
		\caption{Ablation of components in GTAV to Cityscapes.
		}\label{table:ablation}
		\begin{tabular}{cccccc|c}
		\toprule
		ID & RWC & NS(R) & NS(C) & PS & MB & mIoU$_{19}$\\
		\midrule
		baseline &-- &--  &--  &--  &-- & 52.1 \\
		\uppercase\expandafter{\romannumeral1} & \checkmark &--  &--  &--  &-- &  53.3\\
		\uppercase\expandafter{\romannumeral2} & \checkmark &\checkmark  &--  & -- &-- &  53.6\\
		\uppercase\expandafter{\romannumeral3} & \checkmark &\checkmark& \checkmark  &--  &-- &  53.9\\
		\uppercase\expandafter{\romannumeral4} & \checkmark &\checkmark& \checkmark  & \checkmark  &-- & 54.2 \\
		\uppercase\expandafter{\romannumeral5} & \checkmark &\checkmark& \checkmark & \checkmark  & \checkmark & 55.3 \\
		\bottomrule
		\end{tabular}
	\end{table}

\subsection{Ablation Study}
\label{sec:4.3}
In this section, we perform ablation studies to investigate the role of RCCR components, including the region-wise contrastive (RWC) loss, random sampling (NS(R)) and category-wise sampling (NS(C)) for negative samples, sampling strategies for positive samples (PS), and memory bank (MB). 

\paragraph{Effect of each component.}
As shown in Table \ref{table:ablation}, by adding the RWC loss, we boost the strong baseline~\cite{tranheden2020dacs} with an additional $+1.2\%$, achieving $53.3\%$, showing the effectiveness of region-level alignment under different environments. When taking the different sampling strategies, \emph{i.e.,} NS(R) and NS(C), into account, we find the gradual and non-marginal improvement by $0.6\%$, which reveals that combining the category information derived from the label or target segmentation output can lead to a more powerful contrastive learning scheme. Finally, we take the memory bank to store the negative samples created from the last three batches, and this mechanism makes a further improvement by $+1.1\%$. 

\paragraph{Effect of momentum projection heads.} Table \ref{table:ablation2} illustrates the effect of the momentum projection head $F_{proj}$. If removing $F_{proj}$, the results on two benchmarks are $52.1\%$ and $54.8\%$, respectively. If directly using the feature map $z$ from the encoder for the proposed region-wise contrastive (RWC) loss $\mathcal{L}_{cont}(z)$, the improvements are limited with $+0.5\%$ and $+0.2\%$ in two benchmarks. In contrast, with using the embeddings $e$ for computing $\mathcal{L}_{cont}(e)$, we can yield larger improvements by  $+3.2\%$ and $+4.2\%$, respectively, which confirms the effectiveness of our $F_{proj}$. 

\begin{table}[t]
\caption{Effect of momentum projector $F_{proj}$.}
\label{table:ablation2}
\centering
\begin{tabular}{ccc|c|c} \toprule
\thead{Ours \\ (w/o $F_{proj}$)} & $\mathcal{L}_{cont}(z)$& $\mathcal{L}_{cont}(e)$ & \thead{mIoU$_{19}$ \\ (GTAV)}  & \thead{mIoU$_{13}$ \\ (SYN)}\\
\midrule
\checkmark  & -- & --  & 52.1 & 54.8 \\
\checkmark  & \checkmark  & -- & 52.6 & 55.0\\
\checkmark  & -- & \checkmark & 55.3 &59.0 \\
\bottomrule
\end{tabular}
\end{table}

\section{Conclusion}
In this paper, we proposed regional contrastive consistency regularization (RCCR) for domain adaptive semantic segmentation. By maximizing the inter-region differences and minimizing intra-region disagreements, we could effectively manage to keep the regional consistency in a fine-grained manner, \emph{i.e.}, feature space and label space,  regardless of the changing of outdoor environments. Firstly, a region-wise contrastive (RWC) loss, momentum projection heads, and two sampling strategies are proposed to realize the regional consistency. 
Then, we introduce a memory bank mechanism to learn more robust and stable region-wise features under varying environments.  Extensive experimental results demonstrate the state-of-the-art performance of the proposed method. 
\section{Acknowledgement}
Thanks to the the support of National Key Research and Development Program of China (2019YFC1521104), National Natural Science Foundation of China (72192821, 61972157), Shanghai Municipal Science and Technology Major Project  (2021SHZDZX0102), Shanghai Science and Technology Commission (21511101200, 22YF1420300), and Art major project of National Social Science Fund (I8ZD22).

\bibliographystyle{IEEEbib}
\bibliography{icme2022template}

\end{document}